\documentclass[twoside,11pt]{article}

\usepackage{jmlr2e}
\usepackage{amsmath}
\usepackage{natbib}
\usepackage{hyperref}

\newcommand{\argmax}{\mathop{\mathrm{argmax}}}

\ShortHeadings{Distributed Deep Reinforcement Learning}{Distributed Deep Reinforcement Learning}
\firstpageno{1}

\begin{document}

\title{Distributed Deep Reinforcement Learning: An Overview}

\author{\name Mohammad Reza Samsami \email MohammadRezaSamsami76@Gmail.com \\
       \addr Department of Mathematical Sciences\\
       Sharif University of Technology
       \AND
       \name Hossein Alimadad \email Hossein.Alimadad@Gmail.com \\
       \addr Department of Mathematical Sciences\\
       Sharif University of Technology}

\maketitle

\begin{abstract}
 Deep reinforcement learning (DRL) is a very active research area. However, several technical and scientific issues require to be addressed, amongst which we can mention data inefficiency, exploration-exploitation trade-off, and multi-task learning. Therefore, distributed modifications of DRL were introduced; agents that could be run on many machines simultaneously. In this article, we provide a survey of the role of the distributed approaches in DRL. We overview the state of the field, by studying the key research works that have a significant impact on how we can use distributed methods in DRL. We choose to overview these papers, from the perspective of distributed learning, and not the aspect of innovations in reinforcement learning algorithms. Also, we evaluate these methods on different tasks, and compare their performance with each other and with single actor and learner agents.
\end{abstract}


\section{Introduction}
Deep reinforcement learning (DRL) has achieved remarkable success in a range of tasks, from real-world problems \citep{kalashnikov2018qt} to games \citep{silver2016mastering}. Moreover, focus on a large scale has caused to solve some challenging games such as StartCraft 2 \citep{vinyals2017starcraft} and Dota \citep{OpenAI_dota}. While these successes have been considerable, the development has been mostly in single-task performance. Also, DRL heavily relied on trial-and-error and empirical evaluation, and need the agent to explore a lot that makes the time a bottleneck. Consequently, there is a significant demand for distributed approaches to advance this field.
\\
At the same time, progress in distributed systems and algorithms has been fundamental to the recent success of deep learning. Various advancements in developing distributed variants of gradient descent algorithms, and deep learning frameworks \citep{goyal2017accurate, you2017scaling, abadi2016tensorflow} have been invented that scale-up training required for growth in the field. Thus, researchers proposed and developed distributed architectures for DRL empowering agents to learn from experience faster, leverage exploration strategies, and become capable of learning a diverse set of tasks simultaneously.
\\
A novel characteristic of reinforcement learning is that the agent has an active impact on how the data is generated by interacting with its environment and storing experience trajectories. So DRL would utilize distributed methods by generating more data per time unit, in addition to parallelizing gradient descent and other parallelizable computations in training.
\\
This article is an overview of current ongoing research directions by analyzing approaches using distributed methods in DRL and providing a unified view of the state-of-the-art. Our study is not meant to be exhaustive; we just review the research works that have a fundamental contribution to distributed DRL. 
\\
There have also been so many methods to multi-agent reinforcement learning \citep{bansal2017emergent, tampuu2017multiagent, mordatch2018emergence}, in which, there are a number of agents operating within a single shared environment, each of which aims to optimize its long-term return by interacting with the environment and other agents. However, in this article, we review the papers seeking to solve a single-agent problem by exploiting parallel computation and decentralized data generation. Some works apply several distributed frameworks (like MapReduce) in large matrix multiplications \citep{li2011mapreduce}. Nevertheless, they are not applicable to non-linear representations, so we do not review them.
\\
This review paper is organized as follows. In the first step, we provide background on reinforcement learning and distributed learning, and introduce the basic concepts used in the rest of the paper (section 2). Section 3 presents and analyzes the notable approaches, their limitations, and their potential perspectives. We also illustrate a minimal representation for each method for deeper understanding and highlight their commonalities and differences. Finally, in section 4, we conclude by identifying each method's empirical results, achievements, and issues.

\section{Background}
\subsection{Reinforcement Learning}
In a typical reinforcement learning framework \citep{sutton2018reinforcement}, an \textit{agent} interacts sequentially with an \textit{environment} $\mathcal{E}$ to learn what \textit{actions} it requires to take in any given \textit{state} to maximize its \textit{long-term outcomes}. At each time step t, the agent receives state $s_t$ and selects an action $a_t$ from some set of possible actions $\mathcal{A}$ according to a distribution $\pi (.|s_t)$ called the policy (it is agent’s behavior function). In return, the agent receives a reward $r_t \in \mathcal{R}$, and transitions to the next state $s_{t+1}$, according to the reward function $r(s_t, a_t)$ and transition probability $p(s_{t+1}|s_t, a_t)$ respectively. The return
$G_t = \sum_{i=0}^{\infty} \gamma^i r_{t+i}$
is the total collected return from time step t with the discount factor $\gamma$.
\\
By long-term outcome, we mean the \emph{value}, or expected return
$V_{\pi} (s) = \mathbb{E}_{\pi}[G_t | s_t = s]$
by following the policy $\pi$. In practice, we mostly encounter with the value of an action \emph{a} in state \emph{s}:
$Q_{\pi} (s, a) = \mathbb{E}_{\pi}[G_t | s_t = s, a_t = a]$.
\\
To reach the optimal policy, estimating the action-value function $Q_{\pi}$ of the optimal policy is sufficient, $Q_{*}(s, s) = \underset{\pi}{\argmax} \ Q_{\pi}(s, a)$. We could use the Bellman equation \citep{Bellman:1957}, which is a recursive formulation relating our current value to the value we make in the immediate future:
\[ Q_{*} (s, a) = \mathbb{E}[r(s, a)] + \gamma \underset{a'}{\max}\ \mathbb{E}[Q_{*} (s', a')], \]
where $(s, a) \rightarrow s'$ is a random transition. 
\\
In value-based model-free RL algorithms, we represent the action-value function using a function approximator such as a neural network \citep{tsitsiklis1997analysis} with parameters $\theta$:
$Q(s, a) \approx Q(s, a; \theta)$. One example of such a network is deep Q-networks, which aims to directly approximate the optimal action-value function using the Q-learning algorithm. In Q-learning, at every update iteration $i$, the parameters $\theta$ of the action-value function $Q(s, a; \theta)$ are updated to minimize the mean-squared Bellman error, by optimizing the following loss function:

\[ L_i (\phi_i) = \mathbb{E} \big[(r(s, a) + \gamma \underset{a'}{\max}\ Q (s', a' ; \phi_{i-1}) - Q(s, a; \phi_i))^2 \big] \]
\\
In some cases, value functions are complex so they make the learning process slow and unstable. Compared to value-based methods, the policy-based methods model and optimize the policy directly. The policy is usually modeled with a parameterized function $\pi(a|s; \phi)$, and learned by doing gradient ascent on $\mathbb{E}[G]$ \citep{sutton2000policy}. Score-function estimator or REINFORCE \citep{williams1992simple} is one of the most generally-applicable gradient estimators and an example of such a method:
\[ J(\phi) = \nabla_\phi \mathbb{E}[G] \approx \nabla_\phi \log \pi(a|s; \theta)G \]
\\
This estimator is an unbiased estimate of $\nabla_\phi \mathbb{E}[G]$, but usually has high variance. We can subtract a baseline function $b(s)$ from the return $G$ to reduce policy gradient estimation variance while keeping it unbiased deriving
$\nabla_\phi \log \pi(a|s; \theta)(G - b(s))$. It is common to choose the state-value function as a baseline
$V_{\pi}(s) \approx b(s)$
that leads to a much lower variance. We can see
$G - V_{\pi}(s)$ 
as an approximation of the advantage of action \textit{a} in state \textit{s}:
$A(a, s) = Q_{\pi}(s, a) - V_{\pi}(s)$. So it makes sense to learn the value function in addition to the policy, and combine the benefits of both value-based and policy-based approaches. That is what the \textit{Actor-Critic} method does, where actor and critic update the policy parameters and value function parameters, respectively.

\subsubsection{Proximal Policy Optimization}

Proximal policy optimization, PPO \citep{heess2017emergence} is an improvement to Trust region policy optimization, TRPO \citep{schulman2015trust}, a policy-gradient-based method. TRPO could improve the solution to many of the complex continuous learning problems. TRPO's idea emerged from the fact that previously, to achieve the best solutions, many of these kinds of tasks had been solved by precisely adjusting the reward function. However, this approach led to results that would respond poorly to the slightest change to the problem's setting. Thus, the intent was to make the agent capable of acting with the same optimality while receiving a simple reward signal. The approach taken is to confront the agent with a spectrum of challenges in the environment with a growing difficulty. To make the idea beneficial, the information from recent experiences must not disregard early ones. In TRPO, to make the policy gradient robust, updated policy's change from the old one, in terms of Kullback–Leibler divergence, is bound in a trust region \citep{schulman2015trust}. PPO \citep{heess2017emergence}, as an approximation of TRPO, restricts the policy gradient via a KL-regularization term. The regularization term has a coefficient that increases by a scale factor if the policy update goes beyond the upper bound and decreases in the same way if it falls below the lower bound. In contrast with TRPO, PPO eases the use of the framework in an extensively parallelized framework and with RNNs, due to the sole reliance on first-order gradients.

\[
J_{\textrm{PPO}}(\phi) = J(\phi) - \lambda \textrm{KL}[\pi_{\textrm{old}}|\pi_\phi]
\]
\\
Considering the interval $[\beta_{\textrm{lowKLtarget}}, \beta_{\textrm{highKLtarget}}]$ as the constraint on change in KL, in every iteration, if $\textrm{KL}[\pi_{\textrm{old}}|\pi{\phi}] > \beta_{\textrm{highKLtarget}}$, then $\lambda \leftarrow \alpha \lambda$. On the other hand, if $\textrm{KL}[\pi_{\textrm{old}}|\pi{\phi}] < \beta_{\textrm{lowKLtarget}}$, we have $\lambda \leftarrow \lambda / \alpha$.

\subsection{Distributed Learning}

Learning frameworks might seldom take a significant time to conclude, and computation power could not always increase to suit the needs. Parallelization usually comes to the rescue.
Deep learning can get parallelized easily, and many of the distributed learning frameworks exploit this property. The DistBelief framework \citep{dean2012large} introduces a distributed deep learning framework and shows that such approaches can drastically overperform deep learning run on a single machine, even in low parallelization levels. DistBelief benefits from model and data parallelism. Having model parallelism, DistBelief partitions and trains the model among multiple machines, and having data parallelism, it replicates the model, and each replica learns based on a different partition of data. DistBelief method does not assume any high-level synchronicity. \citep{krizhevsky2014one} introduces another way to parallelize training convolutional neural networks with stochastic gradient descent, with a similar structure to the DistBelief method, but instead applying the resulting gradient updates synchronously. Recent RL frameworks mostly take advantage of neural networks for stochastic gradient descent.
Regarding the bottlenecks, \citep{keuper2016distributed} does an in-depth overview of scalability in some supervised learning methods. Due to the requirement of sharing the parameter updates between the nodes, communication overheads emerge. Nevertheless, using larger batches and step sizes is a workaround to the mentioned problem, though it sacrifices the model's accuracy.
Finally, \citep{ben2019demystifying} extensively analyzes concurrency concerns in multiple learning frameworks, an in-depth review of this subject.
A reinforcement learning system consists of the environment, the actor, and the learner, communicating one another. Distributed RL architectures often take advantage of parallelism in either the learner or the actor independently, so it affects the computation cost and efficiency. Nevertheless, the components themselves might have independently parallelized components internally.

\section{Methods and Architectures}
\subsection{GORILA}
\begin{figure}[h!]
	\begin{center}
		\includegraphics[width=0.8\linewidth]{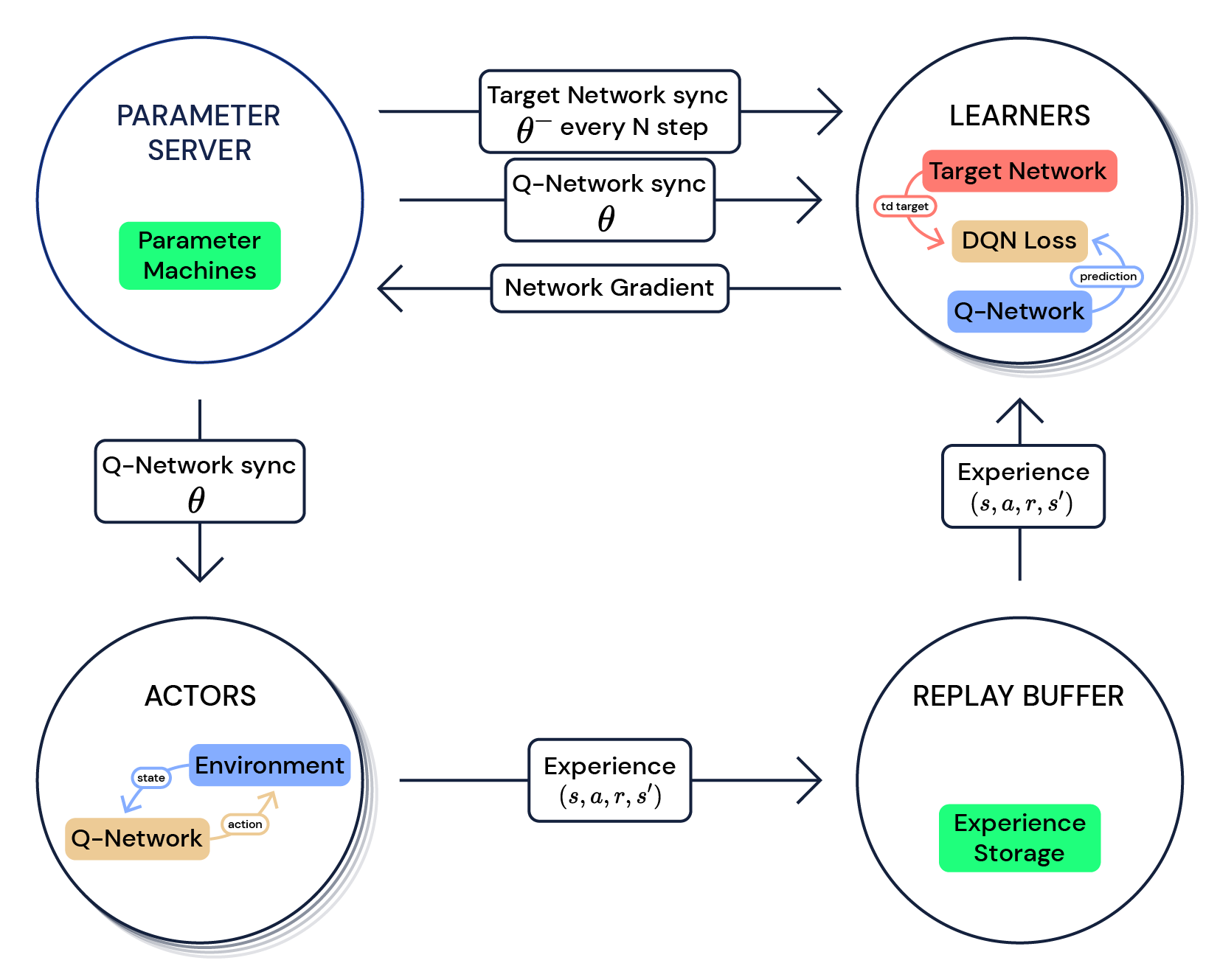}
	\end{center}
	\caption{Overview of GORILA architecture}
	\label{fig:gorila}
\end{figure}
The General Reinforcement Learning Architecture, GORILA \citep{nair2015massively}, is an asynchronous distributed reinforcement learning architecture. Gorila has similarities with DQN [ref] in the algorithm, yet has multiple workers and learners, each running within a single machine, and stores a Q-network inside each. It utilizes the DistBelief method \citep{dean2012large} to compute the stochastic gradient descent.
The Gorila framework architecture consists of 4 main parts. The parameter server, the workers, the learners, and the replay buffer.
The parameter server is the source of truth for the Q-network parameters $\theta$. Workers and learners each store a replica of $\theta$. Therefore, the parameter server tries to keep $\theta$ up-to-date in workers and learners.
Each worker stores the parameters $\theta$, according to which it acts. It is just used for acting and does not partake in the stochastic gradient descent procedure. So it gets synced with the updates to the parameter server. Workers send their experience $(s, a, s', a')$, to the replay memory, placed either on an independent distributed database or inside the workers.
Learners sample experiences from the replay buffer and perform the stochastic gradient descent update. Such in DQN, each learner keeps an "improving" Q-network, with parameters $\theta$ and an assumably "constant" target Q-network, with parameters $\theta^-$. The target network gets updated once in N iterations, such in DQN. The learner's responsibility is to determine the loss from experiences sampled from the replay buffer. Still, the learner does not update the improving Q-network, as opposed to DQN, it sends the resulting gradient to the parameter server. The improving Q-network only gets synced whenever the parameter server updates $\theta$, because there are other learners issuing updates on $\theta$, and the outgoing update is not necessarily valid.
The parameter server contains a DistBelief network. It is responsible for applying the incoming gradients from learners and sending network parameters to the Q-networks in learners and actors, as described above. The parameter server drops "outdated" gradient updates, making the parameters from this source reliable even on process and communication delays and failures.
Several variants of this architecture are categorized as called GORILA. The number of learners and actors could be anything, one or many, and the learner and actors could be "bundled," meaning that there is the same number of learners and actors, each couple placed in one machine. This variant reduces communication overhead, so it has better performance and was used to evaluate the framework.

\subsection{A3C and A2C}
\begin{figure}[h!]
	\begin{center}
		\includegraphics[width=0.8\linewidth]{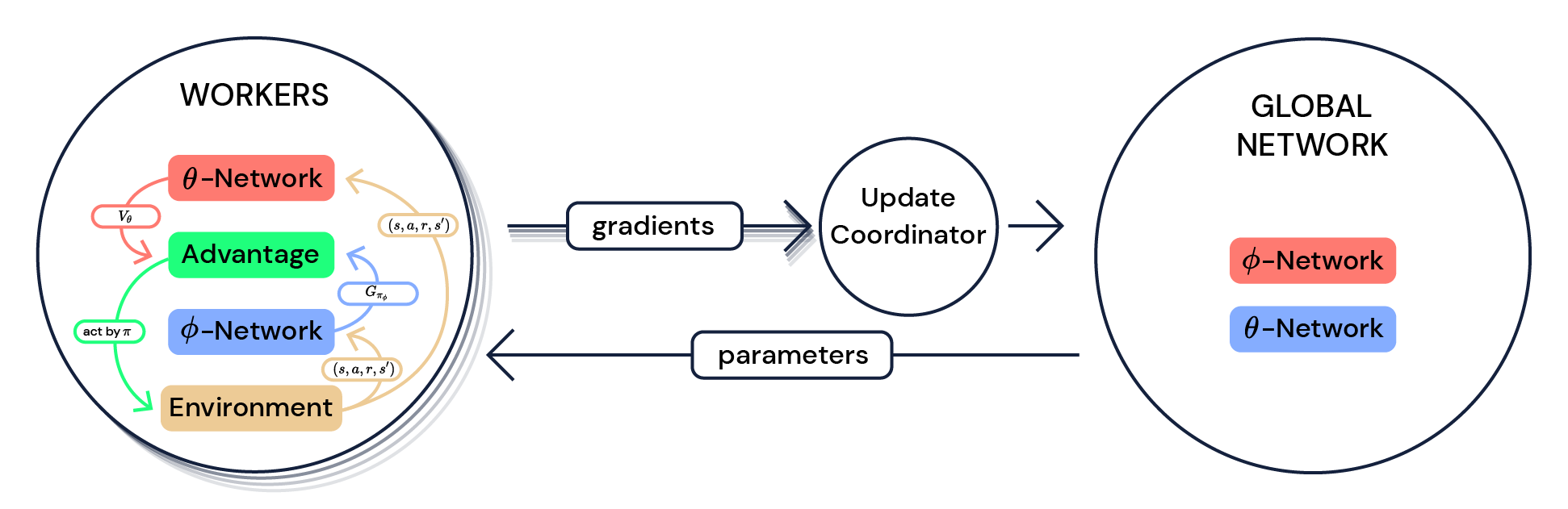}
	\end{center}
	\caption{Overview of A2C architecture}
	\label{fig:a2c}
\end{figure}

In GORILA, the agent collects data in the replay buffer so that the data can be batched or sampled, non-stationarity is decreased, and updates are decorrelated. However, storing experiences in the replay buffer restricts the methods to off-policy algorithms, and requires more computation and memory per interaction. That is why in \textit{Asynchronous Advantage Actor-Critic} \citep{mnih2016asynchronous}, short for \textit{A3C}, multiple agents asynchronously run in parallel to generate data. This approach provides a more practical alternative to experience replay since parallelization also diversifies and decorrelates the data.
\\
\begin{figure}[h!]
	\begin{center}
		\includegraphics[width=0.8\linewidth]{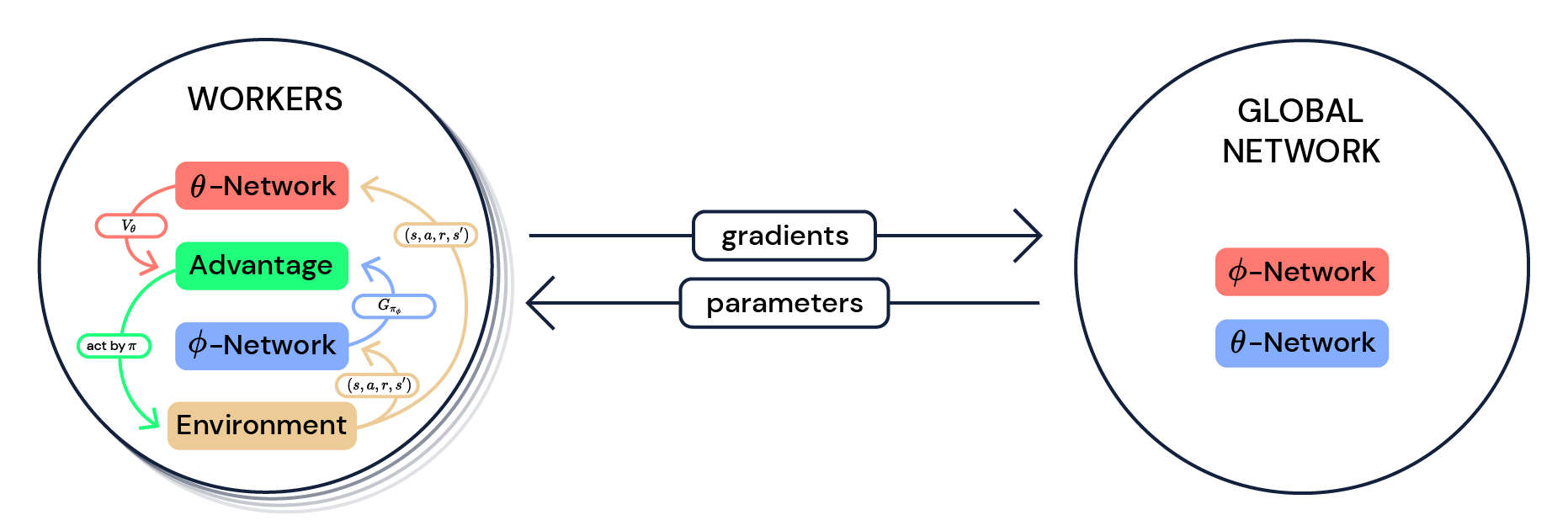}
	\end{center}
	\caption{Overview of A3C architecture}
	\label{fig:a3c}
\end{figure}
\\
In A3C (Figure \ref{fig:a3c}), there is a global network and many worker agents that each has its own parameters. Each of these agents interacts with its copy of the environment simultaneously as the other agents are interacting with their environments, and updates independently of the execution of other agents when they want to update their shared network. It is evident from its name that A3C moves to the actor-critic method from Q-learning compared to GORILA (on-policy methods are more stable than off-policy methods), and couples a DQN with a deep policy network for selecting actions. A3C exploits the multithreading capabilities of standard CPUs while GORILA relies heavily on a massively distributed architecture. Also, keeping the learners on a single machine removes the communication costs of sending gradients and parameters and reduces training time. Moreover, workers can use different exploration policies to explore different parts of the environment. In brief, the outline of every iteration of each worker's program is:
\begin{enumerate}
	\item Reset to the global network.
	\item Interact with its copy of the environment.
	\item Compute value and policy loss, and get gradients.
	\item Update the global network. 
\end{enumerate}\leavevmode
\\
A2C is a synchronous variant of A3C (the first A corresponding asynchronous is removed). In A3C, each worker agent talks to the global parameters independently, so sometimes the workers would perform with varying policies; hence, the accumulated update is not necessarily optimal.  In contrast to A3C, a coordinator in A2C waits for all workers to complete their work before optimizing the global parameters, and then in the next iteration, each worker starts from the same policy. So the synchronized gradient update resolves mentioned inconsistency and works better with large batch sizes yet lacks the independence of workers. An
overview of the architecture is shown in Figure \ref{fig:a2c}. Batched A2C \citep{clemente2017efficient} is a variant of A2C, which at each step, it generates a batch of actions and applies them to a batch of environments. Since game logic and rendering are computationally cheap, batched A2C has high performance in Atari games.

\subsection{IMPALA}
\begin{figure}[h!]
	\begin{center}
		\includegraphics[width=0.8\linewidth]{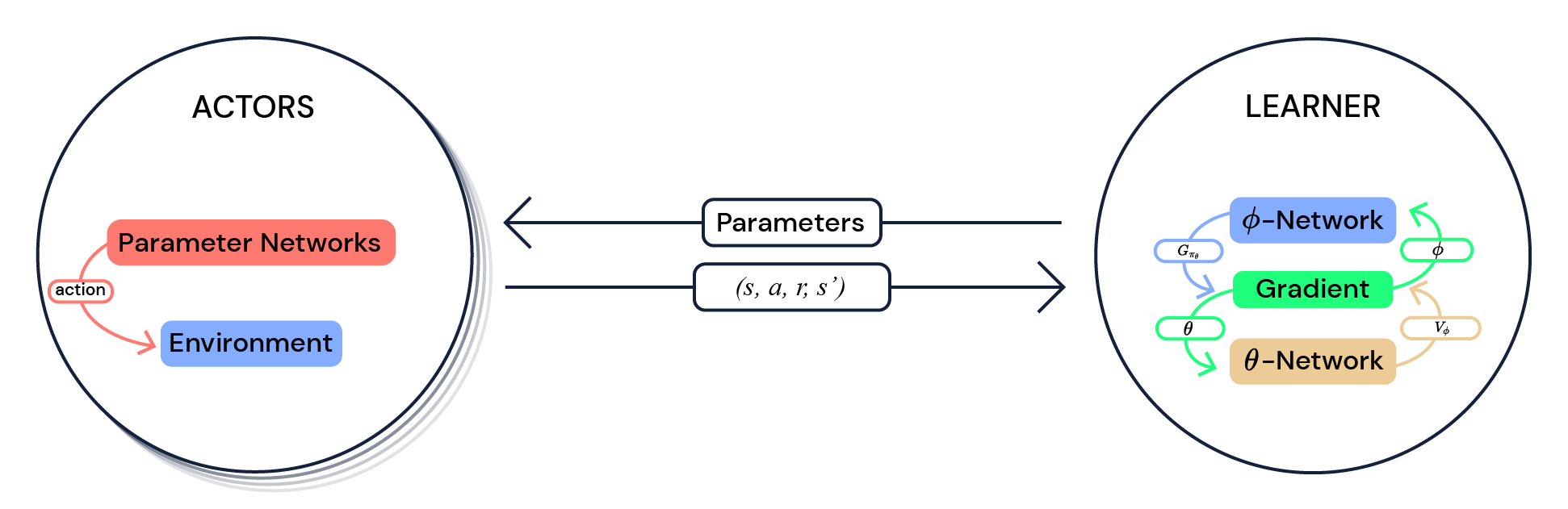}
	\end{center}
	\caption{Overview of IMPALA architecture}
	\label{fig:impala}
\end{figure}

Motivated by progress in distributed deep learning, \citet{espeholt2018impala} proposed \textit{IMPALA} (Importance Weighted Actor-Learner Architecture) to scale-up DRL training to achieve high throughput. Similar to A3C, multiple actors interact with their environments in parallel. On the other hand, IMPALA separates acting from learning, and actors do not calculate gradients. Instead, they only produce trajectories of experience and communicate those data to a central learner, and the learner computes gradients and optimizes both policy and value functions. Each actor renews its parameters with the last version of the policy from the learner periodically (Figure \ref{fig:impala}).
\leavevmode
\\
IMPALA utilizes GPU and distributed deep learning methods to perform updates on mini-batches of data while parallelizing independent operations, and could be implemented on multiple learners learning synchronously between themselves. Decoupling the learning and acting also lets us add many actors to produce many more trajectories per time unit. Diversity in the environment and worker's speed can limit performance in A2C (especially batched A2C), and since the actors do not wait for the learning step in decoupled architecture in IMPALA, the throughput of the system is increased.
\\
At gradient computation time, the actor's policy that has generated experiences might lag behind the learner's policy by updates caused by other actor's trajectories. To correct this gap, we need off-policy revisions; thus, \citet{espeholt2018impala} introduced \textit{V-trace} off-policy correction. Let value and policy functions parameterized by $\theta$ and $\phi$, respectively. Also, suppose a lightly older policy $\mu$ has stored trajectories in the replay buffer. So the n-step V-trace target for $V(s_t;\theta)$ is defined as:
\[ v_t = V(s_t; \theta) + \sum_{i=t}^{t+n-1} \gamma^{i-t} \big(\prod_{j=t}^{i-1} c_j\big)	\rho_i (r_i + \gamma V(s_{i+1};\theta) - V(s_i;\theta)), \]
\\
where
$\rho_i = \min \big(\bar{\rho}, \frac{\pi (a_i|s_i; \phi)}{\mu (a_i|s_i)} \big)$
and
$c_j = \min \big(\bar{c}, \frac{\pi (a_j|s_j; \phi)}{\mu (a_j|s_j)} \big)$ are truncated importance sampling \citep{ionides2008truncated} weights. So the value function parameter is updated in the direction of:
\[ \Delta \theta = (v_t - V(s_t; \theta)) \nabla_{\theta} V(s_t; \theta), \]
\\
and the policy parameters in the direction of the policy gradient:
\[ \Delta \phi = \rho_t \nabla_\phi \log \pi(a_t|s_t; \phi) \big(r_t + \gamma v_{t+1} - V(s_t; \theta) \big) + \nabla_\phi H(\pi_\phi)), \]
\\
where $H(\pi_\phi)$ is an entropy bonus to encourage exploration for preventing premature convergence. More profound details of the algorithm and its analysis can be found in \citet{espeholt2018impala}.

\subsection{Distributed PPO}

Distributed Proximal Policy Optimization, DPPO \citep{heess2017emergence}, is a distributed reinforcement learning framework based on Proximal Policy Optimization, PPO, and improves over A3C in continuous problems. The applied architecture is similar to A2C and pretty straightforward. The architecture consists of a central chief and multiple workers. On every iteration, the chief sends the current parameters to the workers and anticipates the gradients to be calculated by the workers, with the PPO method. When the number of received gradient updates reaches a minimum, the chief applies the gradients' average to the parameter networks for value and policy. Hence, the distributed framework could be considered equivalent to A2C.

\subsection{Ape-X and R2D2}
\begin{figure}[h!]
	\begin{center}
		\includegraphics[width=0.8\linewidth]{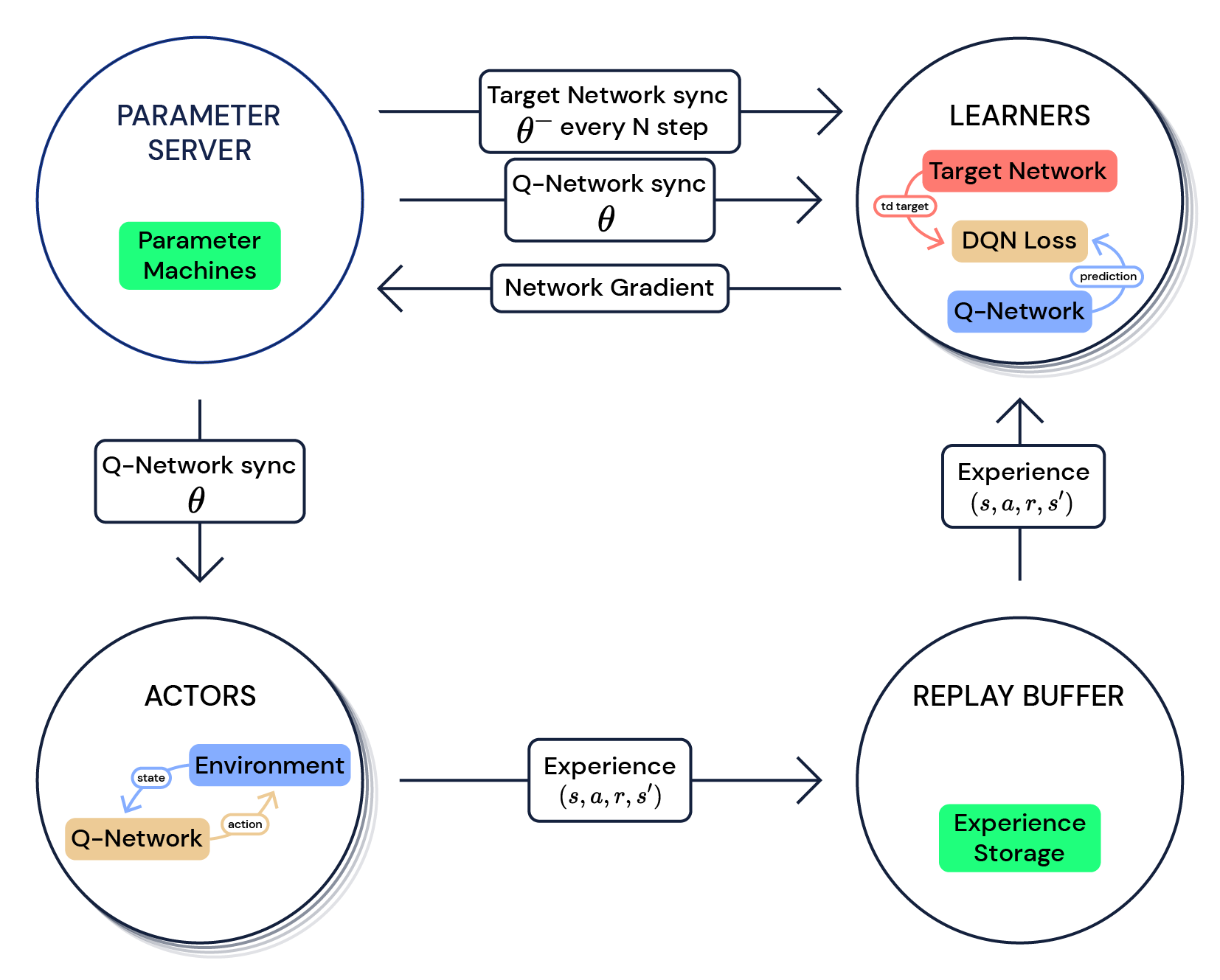}
	\end{center}
	\caption{Overview of Ape-X architecture}
	\label{fig:apex}
\end{figure} \leavevmode
\\
The approach introduced in \citep{horgan2018distributed}, entitled Ape-X, which was used and improved in the method of Recurrent Experience Replay Distributed Reinforcement Learning, R2D2 \citep{kapturowski2018recurrent}, is a value-based framework similar to GORILA from the aspect of the distributed architecture. Ape-X aims to reduce variance and accelerate convergence and benefits from importance sampling and prioritized experience replay. That is, the learners sample the experiences with weights proportional to the difference that they would make. Since the older experiences with more priority are less disregarded, the gradient is less likely to overfit, regarding the latest experiences. So in a sense, prioritized replay aims the same purpose as PPO. Nevertheless, prioritized experience replay alone introduces a bias to the gradient updates; Therefore, importance sampling is responsible for canceling this bias.
The Ape-X framework uses variants of DQN, such as the double Q-learning algorithm \citep{van2016deep}, and Deep Deterministic Policy Gradient (DDPG) \citep{mnih2016asynchronous} multi-step bootstrap targets (Sutton, 1988), as the deep network for evaluation, and uses a dueling network architecture \citep{wang2016dueling} as the function approximator. In Ape-X, parallelism happens more extensively in acting, and the variant with a single learner on a GPU has gained the best empirical performance.
The model network in R2D2 is RNN-based, and like in Ape-X, double Q-learning with multi-step returns is used. R2D2 applies modifications on the model network, and the experience replay stores overlapping sequences of $(s, a, r)$ tuples of experience instead of the default $(s, a, r, s')$ tuples.

\subsection{SEED RL}
\begin{figure}[h!]
	\begin{center}
		\includegraphics[width=0.8\linewidth]{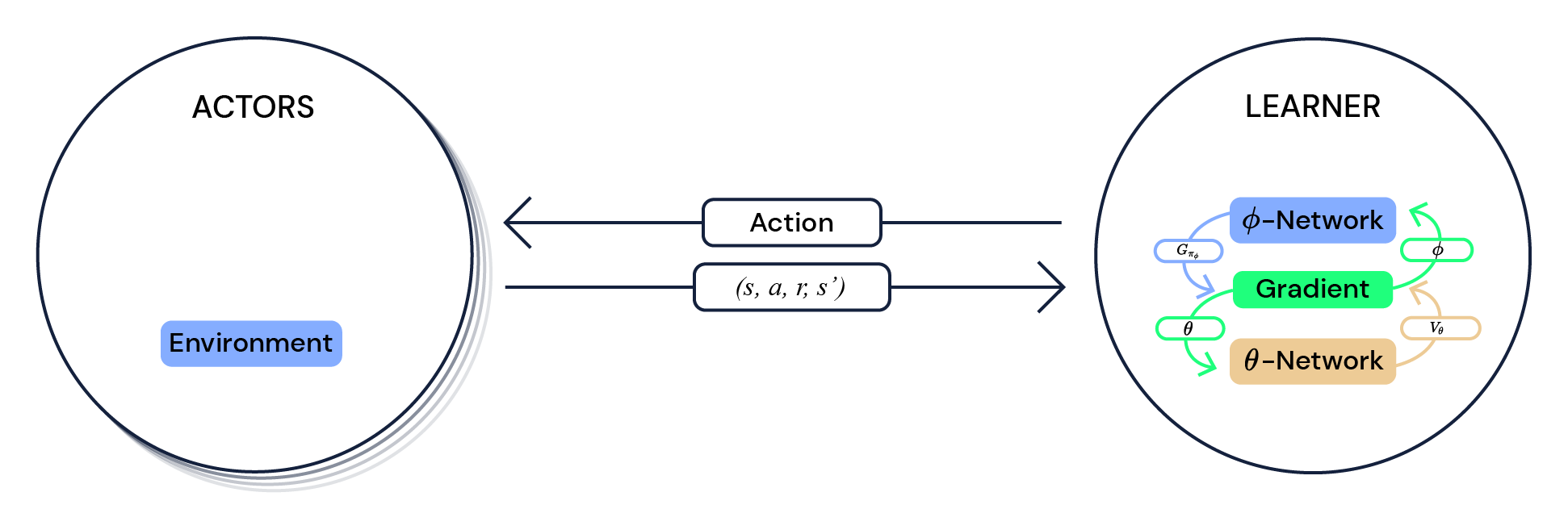}
	\end{center}
	\caption{Overview of SEED RL architecture}
	\label{fig:seed}
\end{figure}

The most recent work analyzed in this paper is \citet{espeholt2019seed} proposing an architecture named SEED RL (Scalable and Efficient Deep-RL), as shown in Figure \ref{fig:seed}, for making the most effective utilization of resources and accelerators, and scaling-up DRL. SEED RL is impressed by IMPALA and invented to solve its drawbacks.
\\
One of IMPALA's drawbacks is that \textbf{actors use CPUs for neural network inference}, which is computationally inefficient than using accelerators \citep{raina2009large}. Seldom, GPUs are used in actors inferences, that limits the number of actors. This issue becomes more critical as models get massive. Moreover, each actor operates two distinct tasks on one machine: model inference and environment rendering. This leads to \textbf{unproductive resource utilization} since these two tasks are not similar. Another issue is that in IMPALA, model parameters and experience trajectories are transferred between actors and learners. Thus, the \textbf{required bandwidth could become a limiting factor}.
\\
A solution to some of these problems is single-machine IMPALA, which avoids CPU-based inferences and does not have bandwidth necessities. However, the number of actors is bounded since they are restricted by resource usage.
\\
The key idea in SEED RL to overcome these drawbacks is that model inference, and trajectory collection is made in learner instead of actors. The whole design is somehow a single-machine structure with remote environments. So we can use batching inference and accelerators such as GPUs and TPUs \citep{jouppi2017datacenter} - AI accelerator application-specific integrated circuit, have multiple environments on the actor, avoid data transfer bottleneck by keeping parameters and states locally in the learner, and there is just a single task in each actor (which is conceptually an environment handler). Additionally, growing the number of parameters does not increase the demand for more actors.
\leavevmode
\\
Nevertheless, a new problem could arise in this model. For each environment iteration, every actor sends its observation to the learner, making a decision, and sending the inferred action back to the actor. So this method can cause latency compared to previous methods in which the inference was not centralized. To decrease this latency, \citet{espeholt2019seed} introduce an efficient library based on \href{http://www.gRPC.io}{gRPC} (a high-performance open-source RPC framework), including asynchronous streaming RPCs (each actor's connection to the learner is kept open). In this framework, a batching module is employed that batches multiple actor inference calls in a well-organized way. Furthermore, the framework utilizes Unix domain sockets in cases where actors can fit on the same machine, so latency and CPU overhead is minimized.
\\
Consequently, SEED RL achieves to a million queries per second on a single machine, the number of actors can be increased to thousands of machines, the learner can be extended to thousands of cores and making it possible to train at millions of frames per second. \citet{espeholt2019seed} adapt two algorithms into the SEED RL to make it successful: V-trace and R2D2. 
\\
Similar to IMPALA, the trajectories are off-policy, so V-trace is applied to compensate for the trajectories. Additionally, R2D2 has many improvements to DQN, especially recurrent distributed replay buffer, enabling RNNs to predict future values based on all past information. However, \citet{espeholt2019seed} implement the replay buffer locally in the learner rather than keeping it distributed and diminish complexity by excluding a load of work. In this setup, the replay buffer is restricted by the learner's memory, but they show it is not a problem for many ranges of tasks.

\section{Results and Conclusion}
\begin{figure}[h!]
	\begin{center}
		\includegraphics[width=0.6\linewidth]{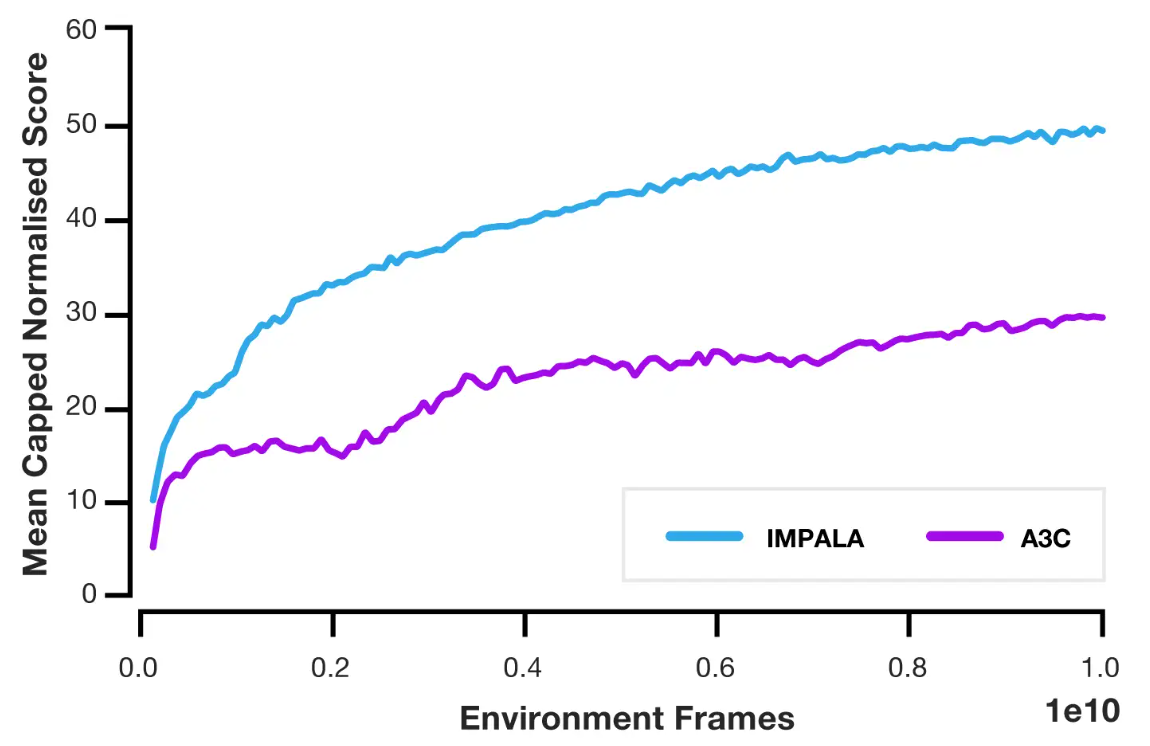}
	\end{center}
	\caption{Performance on DMLab-30 wrt. wall-clock time, from \href{https://deepmind.com/blog/article/impala-scalable-distributed-deeprl-dmlab-30}{DeepMind blog}}
	\label{fig:impalaa3c}
\end{figure}

\begin{figure}[h!]
	\begin{center}
		\includegraphics[width=0.8\linewidth]{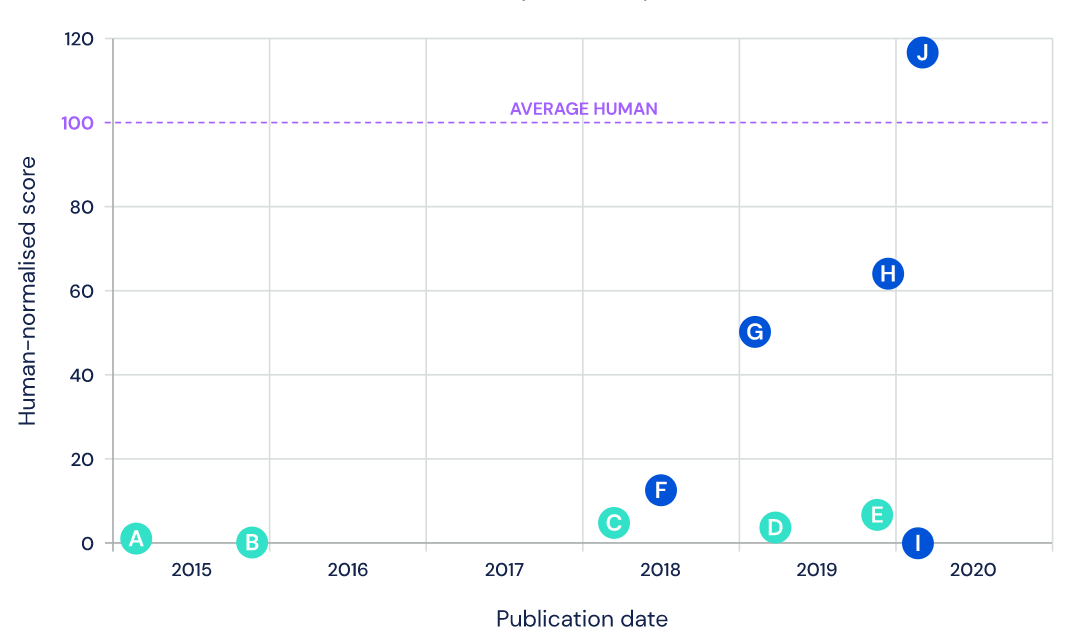}
	\end{center}
	\caption{Agents that use a distributed setup are blue, whereas single-actor agents are teal, from \href{https://deepmind.com/blog/article/Agent57-Outperforming-the-human-Atari-benchmark}{DeepMind blog}}
	\label{fig:agent57}
\end{figure}

We have reviewed distributed versions of standard reinforcement learning algorithms and explained that they could train neural network controllers on numerous domains in a stable way. We showed that stable training of deep neural networks is possible with both value-based and policy-based methods, and off-policy as well as on-policy methods in these frameworks.
\\
\begin{figure}[t]
	\begin{center}
		\includegraphics[width=\linewidth]{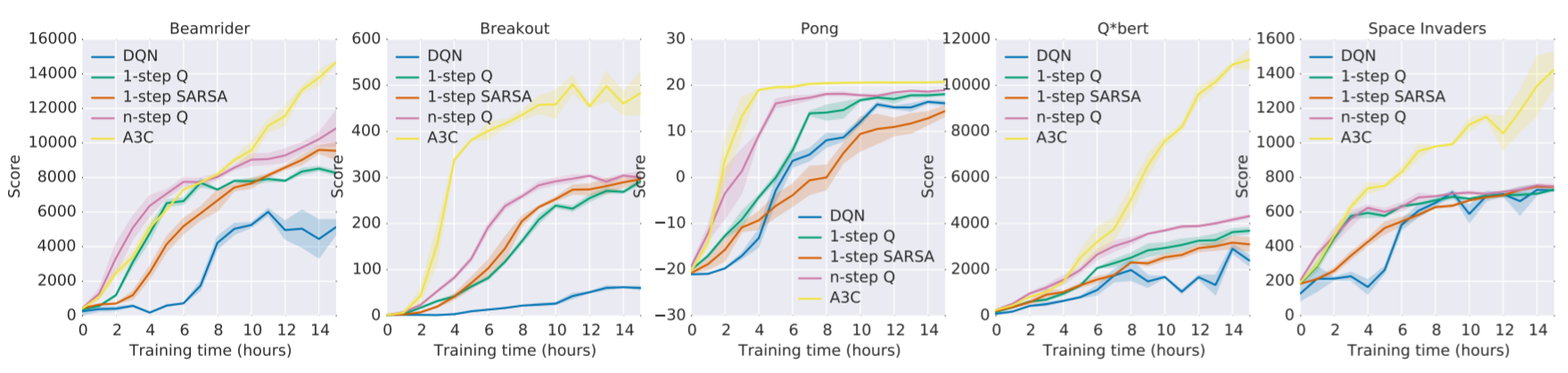}
	\end{center}
	\caption{Learning speed comparison for DQN and asynchronous algorithms, from \citet{mnih2016asynchronous}}
	\label{fig:a3cspeed}
\end{figure}\leavevmode
In general, distributed methods in DRL are more data-efficient, scalable, and faster compared to single-actor methods. Also, they showed positive transfer from training in multi-task settings compared to training in the single-task setting. Furthermore, distributed DRL approaches significantly outperform single DRL methods. For instance, Figure \ref{fig:a3cspeed} compares the learning speed of the DQN algorithm trained on an Nvidia K40 GPU with the A3C trained using 16 CPU cores on five Atari 2600 games.
\\
Distributed DRL methods have been getting improved and evolved through time. Not surprisingly, their overall performance is advancing. For example, Figure \ref{fig:impalaa3c} demonstrates an analysis of IMPALA's performance and A3C performance on DMLab-30.
\leavevmode
\\
Hardware systems and distributed deep learning are growing fast. This considerable growth makes the progress of distributed DRL inevitable. Therefore, we believe deploying distributed DRL techniques would be beneficial and efficient for every researcher with any computational resources. Recently, \citet{badia2020agent57} developed Agent57, the first DRL agent to obtain an above-the-human baseline score on all 57 Atari 2600 games. Different reinforcement learning algorithms and modifications are used Agent57, from distributed architecture to meta-controller for efficient exploration. Figure \ref{fig:agent57} illustrates the performances of some single-actor and some distributed architectures on the top 5$\%$ struggling games in Atari57 (The best one is Agent57). We think it is still an open and increasingly significant research area to scale-up DRL.
\leavevmode

\vskip 0.2in
\bibliography{DRL}

\begin{thebibliography}{35}
\providecommand{\natexlab}[1]{#1}
\providecommand{\url}[1]{\texttt{#1}}
\expandafter\ifx\csname urlstyle\endcsname\relax
  \providecommand{\doi}[1]{doi: #1}\else
  \providecommand{\doi}{doi: \begingroup \urlstyle{rm}\Url}\fi

\bibitem[Abadi et~al.(2016)Abadi, Agarwal, Barham, Brevdo, Chen, Citro,
  Corrado, Davis, Dean, Devin, et~al.]{abadi2016tensorflow}
Mart{\'\i}n Abadi, Ashish Agarwal, Paul Barham, Eugene Brevdo, Zhifeng Chen,
  Craig Citro, Greg~S Corrado, Andy Davis, Jeffrey Dean, Matthieu Devin, et~al.
\newblock Tensorflow: Large-scale machine learning on heterogeneous distributed
  systems.
\newblock \emph{arXiv preprint arXiv:1603.04467}, 2016.

\bibitem[Badia et~al.(2020)Badia, Piot, Kapturowski, Sprechmann, Vitvitskyi,
  Guo, and Blundell]{badia2020agent57}
Adri{\`a}~Puigdom{\`e}nech Badia, Bilal Piot, Steven Kapturowski, Pablo
  Sprechmann, Alex Vitvitskyi, Daniel Guo, and Charles Blundell.
\newblock Agent57: Outperforming the atari human benchmark.
\newblock \emph{arXiv preprint arXiv:2003.13350}, 2020.

\bibitem[Bansal et~al.(2017)Bansal, Pachocki, Sidor, Sutskever, and
  Mordatch]{bansal2017emergent}
Trapit Bansal, Jakub Pachocki, Szymon Sidor, Ilya Sutskever, and Igor Mordatch.
\newblock Emergent complexity via multi-agent competition.
\newblock \emph{arXiv preprint arXiv:1710.03748}, 2017.

\bibitem[Bellman(1957)]{Bellman:1957}
Richard Bellman.
\newblock \emph{Dynamic Programming}.
\newblock Princeton University Press, Princeton, NJ, USA, 1 edition, 1957.

\bibitem[Ben-Nun and Hoefler(2019)]{ben2019demystifying}
Tal Ben-Nun and Torsten Hoefler.
\newblock Demystifying parallel and distributed deep learning: An in-depth
  concurrency analysis.
\newblock \emph{ACM Computing Surveys (CSUR)}, 52\penalty0 (4):\penalty0 1--43,
  2019.

\bibitem[Clemente et~al.(2017)Clemente, Castej{\'o}n, and
  Chandra]{clemente2017efficient}
Alfredo~V Clemente, Humberto~N Castej{\'o}n, and Arjun Chandra.
\newblock Efficient parallel methods for deep reinforcement learning.
\newblock \emph{arXiv preprint arXiv:1705.04862}, 2017.

\bibitem[Dean et~al.(2012)Dean, Corrado, Monga, Chen, Devin, Mao, Ranzato,
  Senior, Tucker, Yang, et~al.]{dean2012large}
Jeffrey Dean, Greg Corrado, Rajat Monga, Kai Chen, Matthieu Devin, Mark Mao,
  Marc'aurelio Ranzato, Andrew Senior, Paul Tucker, Ke~Yang, et~al.
\newblock Large scale distributed deep networks.
\newblock In \emph{Advances in neural information processing systems}, pages
  1223--1231, 2012.

\bibitem[Espeholt et~al.(2018)Espeholt, Soyer, Munos, Simonyan, Mnih, Ward,
  Doron, Firoiu, Harley, Dunning, et~al.]{espeholt2018impala}
Lasse Espeholt, Hubert Soyer, Remi Munos, Karen Simonyan, Volodymir Mnih, Tom
  Ward, Yotam Doron, Vlad Firoiu, Tim Harley, Iain Dunning, et~al.
\newblock Impala: Scalable distributed deep-rl with importance weighted
  actor-learner architectures.
\newblock \emph{arXiv preprint arXiv:1802.01561}, 2018.

\bibitem[Espeholt et~al.(2019)Espeholt, Marinier, Stanczyk, Wang, and
  Michalski]{espeholt2019seed}
Lasse Espeholt, Rapha{\"e}l Marinier, Piotr Stanczyk, Ke~Wang, and Marcin
  Michalski.
\newblock Seed rl: Scalable and efficient deep-rl with accelerated central
  inference.
\newblock \emph{arXiv preprint arXiv:1910.06591}, 2019.

\bibitem[Goyal et~al.(2017)Goyal, Doll{\'a}r, Girshick, Noordhuis, Wesolowski,
  Kyrola, Tulloch, Jia, and He]{goyal2017accurate}
Priya Goyal, Piotr Doll{\'a}r, Ross Girshick, Pieter Noordhuis, Lukasz
  Wesolowski, Aapo Kyrola, Andrew Tulloch, Yangqing Jia, and Kaiming He.
\newblock Accurate, large minibatch sgd: Training imagenet in 1 hour.
\newblock \emph{arXiv preprint arXiv:1706.02677}, 2017.

\bibitem[Heess et~al.(2017)Heess, TB, Sriram, Lemmon, Merel, Wayne, Tassa,
  Erez, Wang, Eslami, et~al.]{heess2017emergence}
Nicolas Heess, Dhruva TB, Srinivasan Sriram, Jay Lemmon, Josh Merel, Greg
  Wayne, Yuval Tassa, Tom Erez, Ziyu Wang, SM~Eslami, et~al.
\newblock Emergence of locomotion behaviours in rich environments.
\newblock \emph{arXiv preprint arXiv:1707.02286}, 2017.

\bibitem[Horgan et~al.(2018)Horgan, Quan, Budden, Barth-Maron, Hessel,
  Van~Hasselt, and Silver]{horgan2018distributed}
Dan Horgan, John Quan, David Budden, Gabriel Barth-Maron, Matteo Hessel, Hado
  Van~Hasselt, and David Silver.
\newblock Distributed prioritized experience replay.
\newblock \emph{arXiv preprint arXiv:1803.00933}, 2018.

\bibitem[Ionides(2008)]{ionides2008truncated}
Edward~L Ionides.
\newblock Truncated importance sampling.
\newblock \emph{Journal of Computational and Graphical Statistics}, 17\penalty0
  (2):\penalty0 295--311, 2008.

\bibitem[Jouppi et~al.(2017)Jouppi, Young, Patil, Patterson, Agrawal, Bajwa,
  Bates, Bhatia, Boden, Borchers, et~al.]{jouppi2017datacenter}
Norman~P Jouppi, Cliff Young, Nishant Patil, David Patterson, Gaurav Agrawal,
  Raminder Bajwa, Sarah Bates, Suresh Bhatia, Nan Boden, Al~Borchers, et~al.
\newblock In-datacenter performance analysis of a tensor processing unit.
\newblock In \emph{Proceedings of the 44th Annual International Symposium on
  Computer Architecture}, pages 1--12, 2017.

\bibitem[Kalashnikov et~al.(2018)Kalashnikov, Irpan, Pastor, Ibarz, Herzog,
  Jang, Quillen, Holly, Kalakrishnan, Vanhoucke, et~al.]{kalashnikov2018qt}
Dmitry Kalashnikov, Alex Irpan, Peter Pastor, Julian Ibarz, Alexander Herzog,
  Eric Jang, Deirdre Quillen, Ethan Holly, Mrinal Kalakrishnan, Vincent
  Vanhoucke, et~al.
\newblock Qt-opt: Scalable deep reinforcement learning for vision-based robotic
  manipulation.
\newblock \emph{arXiv preprint arXiv:1806.10293}, 2018.

\bibitem[Kapturowski et~al.(2018)Kapturowski, Ostrovski, Quan, Munos, and
  Dabney]{kapturowski2018recurrent}
Steven Kapturowski, Georg Ostrovski, John Quan, Remi Munos, and Will Dabney.
\newblock Recurrent experience replay in distributed reinforcement learning.
\newblock In \emph{International conference on learning representations}, 2018.

\bibitem[Keuper and Preundt(2016)]{keuper2016distributed}
Janis Keuper and Franz-Josef Preundt.
\newblock Distributed training of deep neural networks: Theoretical and
  practical limits of parallel scalability.
\newblock In \emph{2016 2nd Workshop on Machine Learning in HPC Environments
  (MLHPC)}, pages 19--26. IEEE, 2016.

\bibitem[Krizhevsky(2014)]{krizhevsky2014one}
Alex Krizhevsky.
\newblock One weird trick for parallelizing convolutional neural networks.
\newblock \emph{arXiv preprint arXiv:1404.5997}, 2014.

\bibitem[Li and Schuurmans(2011)]{li2011mapreduce}
Yuxi Li and Dale Schuurmans.
\newblock Mapreduce for parallel reinforcement learning.
\newblock In \emph{European Workshop on Reinforcement Learning}, pages
  309--320. Springer, 2011.

\bibitem[Mnih et~al.(2016)Mnih, Badia, Mirza, Graves, Lillicrap, Harley,
  Silver, and Kavukcuoglu]{mnih2016asynchronous}
Volodymyr Mnih, Adria~Puigdomenech Badia, Mehdi Mirza, Alex Graves, Timothy
  Lillicrap, Tim Harley, David Silver, and Koray Kavukcuoglu.
\newblock Asynchronous methods for deep reinforcement learning.
\newblock In \emph{International conference on machine learning}, pages
  1928--1937, 2016.

\bibitem[Mordatch and Abbeel(2018)]{mordatch2018emergence}
Igor Mordatch and Pieter Abbeel.
\newblock Emergence of grounded compositional language in multi-agent
  populations.
\newblock In \emph{Thirty-Second AAAI Conference on Artificial Intelligence},
  2018.

\bibitem[Nair et~al.(2015)Nair, Srinivasan, Blackwell, Alcicek, Fearon,
  De~Maria, Panneershelvam, Suleyman, Beattie, Petersen,
  et~al.]{nair2015massively}
Arun Nair, Praveen Srinivasan, Sam Blackwell, Cagdas Alcicek, Rory Fearon,
  Alessandro De~Maria, Vedavyas Panneershelvam, Mustafa Suleyman, Charles
  Beattie, Stig Petersen, et~al.
\newblock Massively parallel methods for deep reinforcement learning.
\newblock \emph{arXiv preprint arXiv:1507.04296}, 2015.

\bibitem[OpenAI()]{OpenAI_dota}
OpenAI.
\newblock Openai five.
\newblock \url{https://blog.openai.com/openai-five/}.

\bibitem[Raina et~al.(2009)Raina, Madhavan, and Ng]{raina2009large}
Rajat Raina, Anand Madhavan, and Andrew~Y Ng.
\newblock Large-scale deep unsupervised learning using graphics processors.
\newblock In \emph{Proceedings of the 26th annual international conference on
  machine learning}, pages 873--880, 2009.

\bibitem[Schulman et~al.(2015)Schulman, Levine, Abbeel, Jordan, and
  Moritz]{schulman2015trust}
John Schulman, Sergey Levine, Pieter Abbeel, Michael Jordan, and Philipp
  Moritz.
\newblock Trust region policy optimization.
\newblock In \emph{International conference on machine learning}, pages
  1889--1897, 2015.

\bibitem[Silver et~al.(2016)Silver, Huang, Maddison, Guez, Sifre, Van
  Den~Driessche, Schrittwieser, Antonoglou, Panneershelvam, Lanctot,
  et~al.]{silver2016mastering}
David Silver, Aja Huang, Chris~J Maddison, Arthur Guez, Laurent Sifre, George
  Van Den~Driessche, Julian Schrittwieser, Ioannis Antonoglou, Veda
  Panneershelvam, Marc Lanctot, et~al.
\newblock Mastering the game of go with deep neural networks and tree search.
\newblock \emph{nature}, 529\penalty0 (7587):\penalty0 484--489, 2016.

\bibitem[Sutton and Barto(2018)]{sutton2018reinforcement}
Richard~S Sutton and Andrew~G Barto.
\newblock \emph{Reinforcement learning: An introduction}.
\newblock 2018.

\bibitem[Sutton et~al.(2000)Sutton, McAllester, Singh, and
  Mansour]{sutton2000policy}
Richard~S Sutton, David~A McAllester, Satinder~P Singh, and Yishay Mansour.
\newblock Policy gradient methods for reinforcement learning with function
  approximation.
\newblock In \emph{Advances in neural information processing systems}, pages
  1057--1063, 2000.

\bibitem[Tampuu et~al.(2017)Tampuu, Matiisen, Kodelja, Kuzovkin, Korjus, Aru,
  Aru, and Vicente]{tampuu2017multiagent}
Ardi Tampuu, Tambet Matiisen, Dorian Kodelja, Ilya Kuzovkin, Kristjan Korjus,
  Juhan Aru, Jaan Aru, and Raul Vicente.
\newblock Multiagent cooperation and competition with deep reinforcement
  learning.
\newblock \emph{PloS one}, 12\penalty0 (4):\penalty0 e0172395, 2017.

\bibitem[Tsitsiklis and Van~Roy(1997)]{tsitsiklis1997analysis}
John~N Tsitsiklis and Benjamin Van~Roy.
\newblock Analysis of temporal-diffference learning with function
  approximation.
\newblock In \emph{Advances in neural information processing systems}, pages
  1075--1081, 1997.

\bibitem[Van~Hasselt et~al.(2016)Van~Hasselt, Guez, and Silver]{van2016deep}
Hado Van~Hasselt, Arthur Guez, and David Silver.
\newblock Deep reinforcement learning with double q-learning.
\newblock In \emph{Thirtieth AAAI conference on artificial intelligence}, 2016.

\bibitem[Vinyals et~al.(2017)Vinyals, Ewalds, Bartunov, Georgiev, Vezhnevets,
  Yeo, Makhzani, K{\"u}ttler, Agapiou, Schrittwieser,
  et~al.]{vinyals2017starcraft}
Oriol Vinyals, Timo Ewalds, Sergey Bartunov, Petko Georgiev, Alexander~Sasha
  Vezhnevets, Michelle Yeo, Alireza Makhzani, Heinrich K{\"u}ttler, John
  Agapiou, Julian Schrittwieser, et~al.
\newblock Starcraft ii: A new challenge for reinforcement learning.
\newblock \emph{arXiv preprint arXiv:1708.04782}, 2017.

\bibitem[Wang et~al.(2016)Wang, Schaul, Hessel, Hasselt, Lanctot, and
  Freitas]{wang2016dueling}
Ziyu Wang, Tom Schaul, Matteo Hessel, Hado Hasselt, Marc Lanctot, and Nando
  Freitas.
\newblock Dueling network architectures for deep reinforcement learning.
\newblock In \emph{International conference on machine learning}, pages
  1995--2003, 2016.

\bibitem[Williams(1992)]{williams1992simple}
Ronald~J Williams.
\newblock Simple statistical gradient-following algorithms for connectionist
  reinforcement learning.
\newblock \emph{Machine learning}, 8\penalty0 (3-4):\penalty0 229--256, 1992.

\bibitem[You et~al.(2017)You, Gitman, and Ginsburg]{you2017scaling}
Yang You, Igor Gitman, and Boris Ginsburg.
\newblock Scaling sgd batch size to 32k for imagenet training.
\newblock \emph{arXiv preprint arXiv:1708.03888}, 6, 2017.

\end{thebibliography}

\end{document}